\documentclass{article}

\usepackage{times}
\usepackage{graphicx} 
\usepackage{subfigure} 

\usepackage{natbib}

\usepackage{algorithm}
\usepackage{algorithmic}
\usepackage{soul}

\newcommand{\err}{\epsilon}

\newcommand{\old}[1]{}

\usepackage[accepted]{icml2016}

\icmltitlerunning{Conceptual Compression}

\begin{document} 

\twocolumn[
\icmltitle{Towards Conceptual Compression}

\icmlauthor{Karol Gregor}{karolg@google.com}
\icmlauthor{Frederic Besse}{fbesse@google.com}
\icmlauthor{Danilo Jimenez Rezende}{danilor@google.com}
\icmlauthor{Ivo Danihelka}{danihelka@google.com}
\icmlauthor{Daan Wierstra}{wierstra@google.com}
\icmladdress{Google DeepMind, London, United Kingdom}

\icmlkeywords{Deep Learning, Generative Models}

\vskip 0.3in
]


\begin{abstract} 
We introduce a simple recurrent variational auto-encoder architecture that significantly improves image modeling. 
The system represents the state-of-the-art in latent variable models for both the ImageNet and Omniglot datasets. 
We show that it naturally separates global conceptual information from lower level details, thus addressing one of the fundamentally desired properties of unsupervised learning. 
Furthermore, the possibility of restricting ourselves to storing only global information about an image allows us to achieve high quality `conceptual compression'.
\end{abstract} 



\section{Introduction}
\label{Introduction}

Images contain a large amount of information that is a priori stored independently in the pixels. In the semi-supervised learning regime where a large number of images is available but only a small number of labels, one would like to leverage this information to create representations that allow for better (and especially faster) generalization. Intuitively one expects such representations to explicitly extract global conceptual aspects of an image.

\begin{figure}[h] 
\vspace{-.0cm}
\begin{center}
\begin{minipage}{0.47\textwidth}
\includegraphics[width=0.24\textwidth]{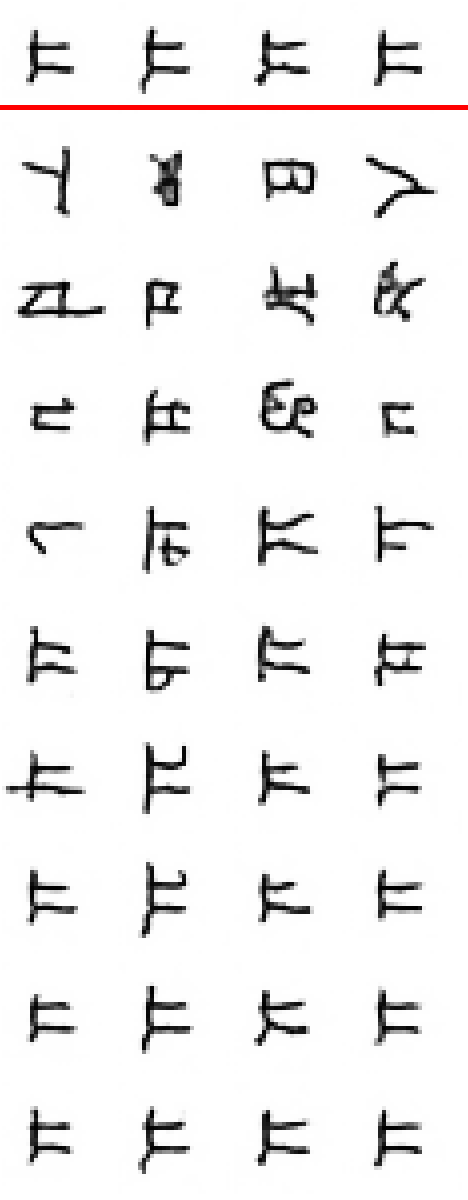}
\includegraphics[width=0.24\textwidth]{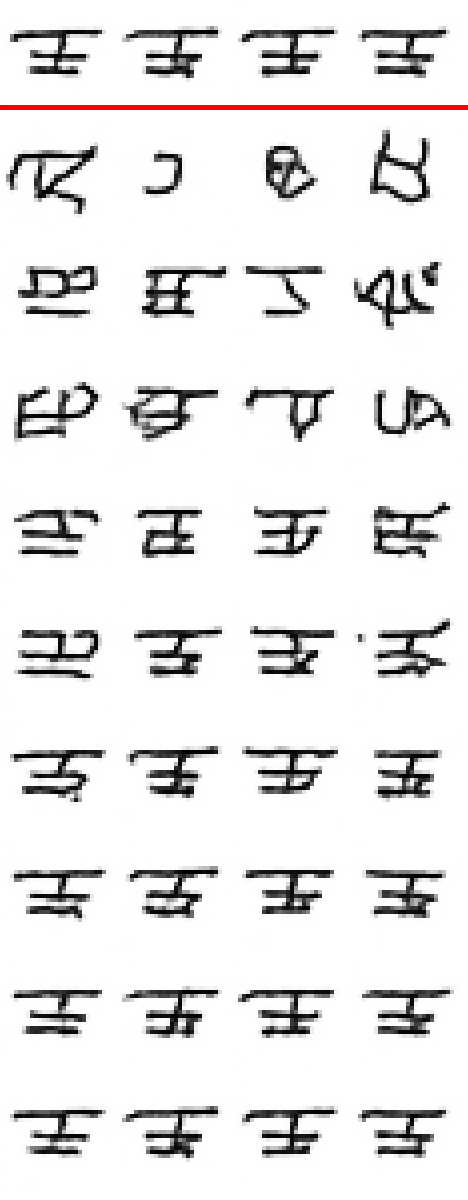}
\includegraphics[width=0.24\textwidth]{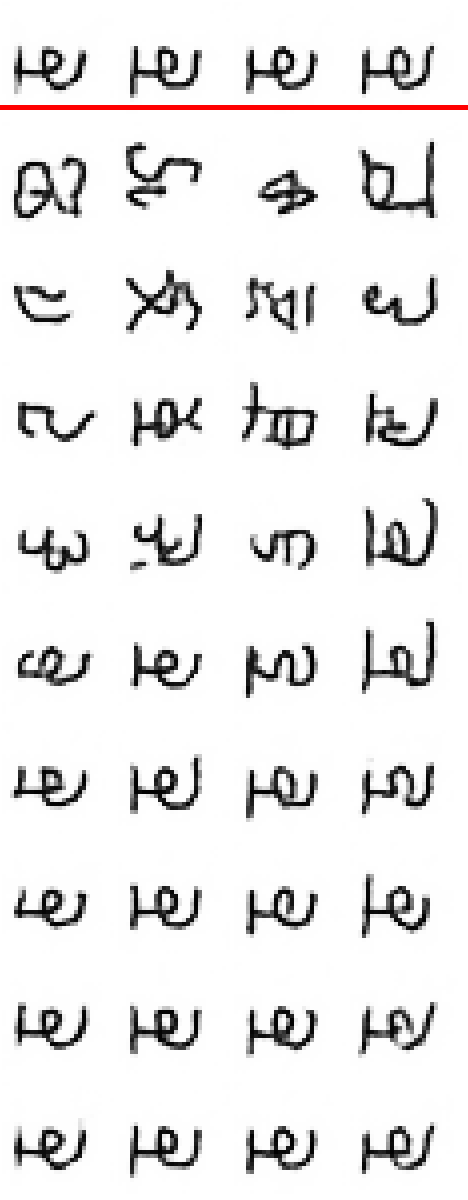}
\includegraphics[width=0.24\textwidth]{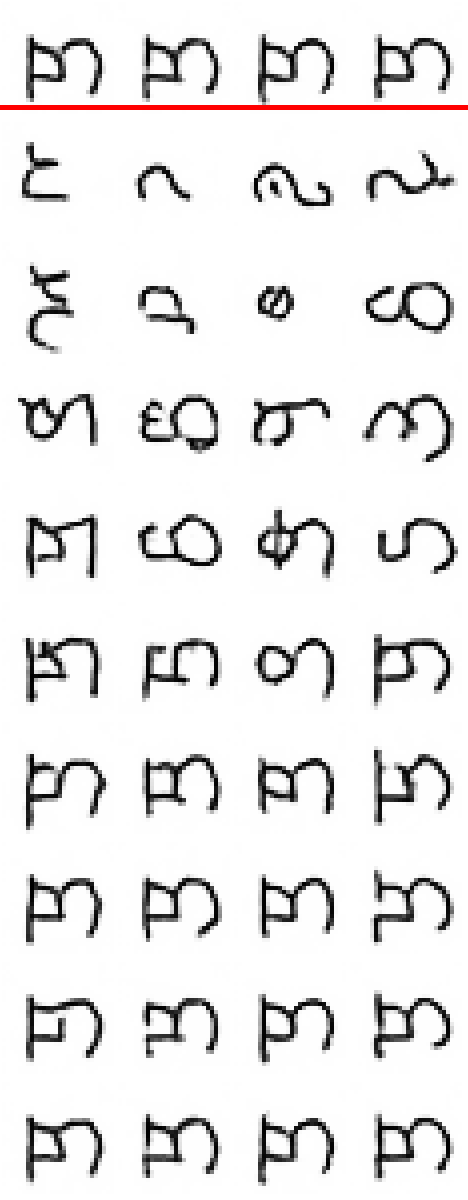}
\end{minipage}
\end{center}
\caption{\textbf{Conceptual Compression [Omniglot]}.
The top row shows full reconstructions from the model. The subsequent rows were obtained by storing the first $t$ groups of latent variables and generating the remaining ones from the model ($t=1, 4, 7, 10, 13, 16, 19, 22, 25, 28$ are shown, out of a total of $30$ steps, from top to bottom). Each group of four columns shows different samples at a given compression level. We see that variations in later samples lie in small details, such as the precise placement of strokes. Reducing the number of stored bits tends to preserve the overall shape, but increases the symbol variation. Eventually a varied set of symbols are generated. Nevertheless even in the first row there is a clear difference between variations produced from a given symbol and those between different symbols. 
}
\label{fig:informationContentOmniglot}
\vspace{-.0cm}
\end{figure}

\begin{figure}[h] 
\vspace{-.0cm}
\begin{center}
\begin{minipage}{0.5\textwidth}
\includegraphics[width=0.95\textwidth]{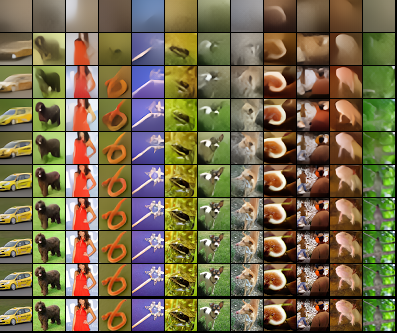}
\end{minipage}
\end{center}
\caption{\textbf{Conceptual Compression [ImageNet]} 
Analogous to Figure \ref{fig:informationContentOmniglot} but applied to natural images. Originals are placed on the bottom to compare more easily to the final reconstructions, which are nearly perfect. Here the latent variables were generated with zero variance. Iterations $t = 2, 4, 6, 8, 10, 14, 18, 25, 32$ of the model with $32$ steps are shown.
}
\label{fig:informationContent}
\vspace{-.0cm}
\end{figure}

In this paper we propose a method that is able to transform an image into a progression of increasingly detailed representations, ranging from global conceptual aspects to low level details (see Figures \ref{fig:informationContentOmniglot} \& \ref{fig:informationContent}). At the same time, our model greatly improves latent variable image modeling compared to earlier implementations of deep variational auto-encoders \citep{kingma2013auto, rezende2014stochastic, gregor2013deep}. Furthermore, it has the advantage of being a simple homogeneous architecture not requiring complex design choices, which is similar to the recurrent structure of DRAW (\citealp{gregor2015draw}). Last, it provides an important insight into building good variational auto-encoder models of images: the use of multiple layers of stochastic variables that are all `close' to the pixels significantly improves performance. 

The system's ability to stratify information enables it to perform high quality lossy compression, by storing only a subset of latent variables, starting with the high level ones, and generating the remainder during decompression (see Figure~\ref{fig:compression}).

Currently the ultimate arbiter of lossy compression remains human evaluation.  Other simple measures such as the L2 distance between compressed and original images are inappropriate -- for example if a particular generated grass texture is sharp, but different from the one in the original image, it will yield a large L2 distance yet should, at the same time, be considered conceptually close to the original.

Achieving good lossy compression while storing only high level latent variables would imply that representations learned at a high level contain information similar to that used by humans to judge images. As humans outperform the best machines at learning abstract representations, human evaluation of lossy compression obtained by these generative models constitutes a reasonable test of the quality of representations learned by these models.

In the following we discuss variational auto-encoders and compression in more detail, present the algorithm and demonstrate the results on generation quality and compression.

\subsection{Variational Auto-Encoders}

Numerous techniques exist for unsupervised learning in deep networks, e.g. sparse auto-encoders and sparse coding \citep{kavukcuoglu2010learning, le2013building}, denoising auto-encoders \citep{vincent2010stacked}, deconvolutional networks \citep{zeiler2010deconvolutional}, restricted Boltzmann machines \citep{hinton2006reducing}, deep Boltzmann machines \cite{salakhutdinov2009deep}, generative adversarial networks \citep{goodfellow2014generative} and variational auto-encoders \citep{kingma2013auto, rezende2014stochastic, gregor2013deep}.  

In this paper we focus on the class of models in the variational auto-encoding framework. Since we are also interested in compression, we present them from an information-theoretic perspective. Variational auto-encoders typically consist of two neural networks: one that generates samples from latent variables (`imagination'), and one that infers latent variables from observations (`recognition'). The two networks share the latent variables. Intuitively speaking one might think of these variables as specifying, for a given image, at different levels of abstraction, whether a particular object such as a cat or a dog is present in the input, or perhaps what the exact position and intensity of an edge at a given location might be.

During the recognition phase the network acquires information about the input and stores it in the latent variables, reducing their uncertainty. For example, at first not knowing whether a cat or a dog is present in the image, the network observes the input and becomes nearly certain that it is a cat. The reduction in uncertainty is quantitatively equal to the amount of information the network acquired about the input. During generation the network starts with uncertain latent variables and selects their values from a prior distribution that specifies this uncertainty (e.g. it chooses a dog). Different choices will produce different samples.



Variational auto-encoders provide a natural framework for unsupervised learning -- we can build networks with layers of stochastic variables and expect that, after learning, the representations become increasingly more abstract for higher levels of the hierarchy. The questions then are: can such a framework indeed discover such representations both in principle and in practice, are such networks powerful enough for modeling real data, and what techniques one needs to make it work well.


\subsection{Conceptual Compression}

Variational auto-encoders can not only be used for representation learning but also for compression. The training objective of variational auto-encoders is to compress the total amount of information needed to encode the input. They achieve this by using information-carrying latent variables that express what, before compression, was encoded using a larger amount of information in the input. The information in the layers and the remaining information in the input can be encoded in practice as explained later in this paper. 

The amount of lossless compression one is able to achieve is bounded by the underlying entropy of the image distribution. Additionally, most image information as measured in bits is contained in the fine details of the image. Thus we might reasonably expect that lossless compression will never improve by more than a factor of two in comparison to current performance. 




\begin{figure}[t]
\vspace{-.0cm}
\begin{center}
\begin{minipage}{0.5\textwidth}
\includegraphics[width=0.99\textwidth]{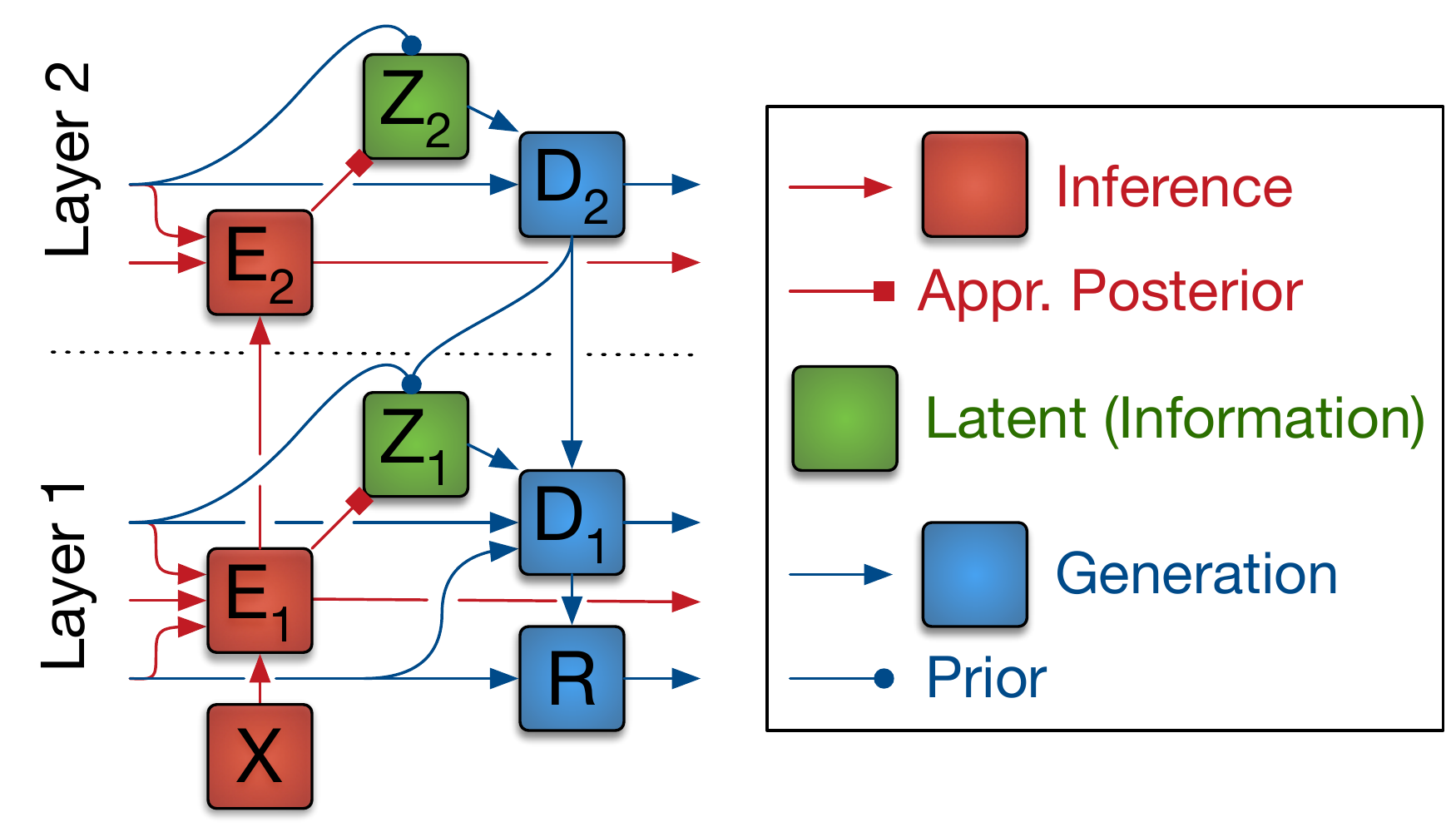}
\caption{\textbf{Schematic depiction of one time slice in convolutional DRAW}. $X$ and $R$ denote input and reconstruction respectively.
}
\vspace{-.5cm}
\label{diagram}
\end{minipage}
\end{center}
\end{figure}

Lossy compression, on the other hand, holds much more potential for improvement. In this case we want to compress an image by a certain amount, allowing some information loss, while maximizing both quality and similarity to the original image. As an example, at a low level of compression (close to lossless compression), we could start by reducing pixel precision, e.g. from 8 bits to 7 bits. Then, as in JPEG, we could express a local 8x8 neighborhood in a discrete cosine transform basis and store only the most significant components. This way, instead of introducing quantization artifacts in the image that would appear if we kept decreasing pixel precision, we preserve higher level structures but to a lower level of precision. However, what can we do beyond that as we keep pushing the compression? We would like to preserve the most important aspects of the image. What determines what is important?


Let us imagine that we are compressing images of cats and dogs and would like to compress an image down to one bit. What would that bit be? One would imagine that it should represent whether the image contains either a cat or a dog. How would we then get an image out of this single bit? If we have a good generative model, we can simply generate the entire image from this one latent variable, an image of a cat if the bit corresponds to `cat', and an image of a dog otherwise. Now let us imagine that instead of compressing to one bit we wanted to compress down to ten bits. Now we can store the most important properties of the animal as well -- e.g. its type, color, and basic pose. The rest would be `filled in' by the generative model that is conditioned on this information. If we increase the number of bits further we can preserve more and more about the image, while generating the fine details such as hair, or the exact pattern of the floor, etc. Most bits are in fact about such low level details. We call this kind of compression -- compressing by giving priority to higher levels of representation and generating the remainder -- `conceptual compression'. We suggest that this should be the ultimate objective of lossy compression. 

Importantly, if we solve deep representation learning with latent variable generative models that generate high quality samples, we achieve the objective of lossy compression mentioned above. We can see this as follows. Assume that the network has learned a hierarchy of progressively more abstract representations. Then, to get different levels of compression, we can store only the corresponding number of topmost layers and generate the rest. By solving unsupervised deep learning, the network would order information according to its importance and store it with that priority.

While the ultimate goal of unsupervised learning remains elusive, we make a step in this direction, and show that our network learns to order information from a rather global level to precise details in images, without being hand-engineered to do this explicitly, as illustrated in Figures~\ref{fig:informationContentOmniglot} and~\ref{fig:informationContent}. This information separation already allows us to achieve better compression quality than JPEG and JPEG2000 as shown in Figure~\ref{fig:compression}. While we are not bound by the same constraints as these algorithms, such as speed and memory, these results demonstrate the potential of this method, which will get better as latent variable generative models improve.


%
%

\subsection{The Importance of Recurrent Feedback}

What are the challenges involved in turning latent variable models into state-of-the-art generative models of images? Many successful vision architectures (e.g. \citealp{simonyan2014very}) have highly over-complete representations that contain many more neurons in hidden layers than pixels. These representations need to be combined to get a very sharp distribution at the pixel level if the pixels are modeled independently. This distribution corresponds to salt and pepper noise which is not present in natural images to a perceptible level. This poses a major challenge.

After experimenting with deep variational auto-encoders we concluded that it was exceedingly difficult to obtain satisfactory results with a single computational pass through the network. Instead we propose that the network needs the ability to correct itself over a number of time steps. Thus, sharp reconstructions should not be a property of high-precision values in the network, but should rather be the result of an iterative feedback mechanism that is robust to network parameter change. 

Such a mechanism is provided by the DRAW algorithm \citep{gregor2015draw}, which is a recurrent type of variational auto-encoder. At each time step, DRAW maintains a provisionary reconstruction, takes in information about a given image, stores it in latent variables and updates the reconstruction. Keeping track of the reconstruction aids the iterative feedback mechanism which is learned by back-propagation. Computation is both deep -- in iterations -- and close to the pixels. 

We introduce convolutional DRAW. It features convolutions, latent prior modeling, a Gaussian input distribution (for natural images) and, in some experiments, a multi-layer architecture. However, it does not use an explicit attentional mechanism. We note that even the single-layer version is already a deep generative model which can decide to process higher level information first before focusing on details, as we demonstrate to some degree. We also experiment with making convolutional DRAW hierarchical in a similar way that we would build conventional deep variational auto-encoders \citep{gregor2013deep} --  stacking more layers of latent and deterministic variables. We believe that the recurrence is important not just for accurate pixel reconstructions, but also at higher levels. For example, when the network decides to generate edges at different locations, it needs to make sure that they are aligned. It is hard to imagine this happening in a single computational pass through the network. Similarly at higher levels, when it decides to generate objects, they need to be generated with the right relationship to one another. And finally at the scene level, one probably does not generate entire scenes at once, but rather one step at a time.

\subsection{Comparison to Non-variational Models}

Let us relate this discussion to two other families of generative models, specifically generative adversarial networks (GANs; \citet{goodfellow2014generative}) and auto-regressive pixel models. GANs have been demonstrated to be able to generate realistic looking images \cite{denton2015deep, radford2015unsupervised}, with properly aligned edges, using a simple feedforward generative network \cite{radford2015unsupervised}. GANs also contain two networks -- a generative network that is the same as in variational auto-encoders, and a classification network. The classification network is presented with both real and model-generated images and tries to classify them according to their true nature -- real or model-generated. The generative network gets gradients from the classification network, changing its weights in an attempt to make the generated images be judged as real ones by the classification network. This makes the generation network produce realistic images that `fool' the classification network. It needs to produce a wide diversity of images, not just one realistic image, because if it produced only one (or a small number of them), the classification network would classify that image as generated and others as realistic, and be almost always correct. This actually happens in practice, and one has to apply a variety of techniques, e.g. as in \cite{radford2015unsupervised}, to obtain sufficient image diversity. However the extent of GANs' sampling diversity is unknown and currently there is no satisfactory measure for it. So while a given network doesn't produce just one image, it is possible that it produces only a tiny subset of possible realistic images, as it simply competes with the power of the classifier. For example if it generates a sharp image, it is unclear whether the system is also capable of generating its translated version, or simply a slightly distorted version.


Finally there is another way to get low uncertainty at the pixel level: instead of predicting pixels independently given the latents, we can decide not to use latents and iterate sequentially over the pixels, predicting a given pixel from the previous ones (from top left to bottom right) in an autoregressive fashion \citep{bengio1999modeling, graves2009offline, larochelle2011neural, gregor2011learning, oord2016pixel}. This is as `close' to the pixels as one can possibly be, and furthermore the procedure is purely deterministic. The disadvantage is conceptual -- the information and decisions are not done at a conceptual level but at the pixel level. For example when generating cats vs dogs the decisions at the first set of pixels (top left of the image) will contain no information about a hypothetical cat or dog. But as we get to the region where these objects are, we start choosing pixels that will slowly tip the probability of generating a cat vs a dog one way or the other. As we start generating an ear, it will more likely be a cat's or a dog's and so on. 
However this pixel level approach and our approach are orthogonal and can be easily combined, for example by feeding the output of convolutional DRAW into the conditional computation of a pixel level model. In this paper we study the latent variable approach and make the pixels independent given the latents.


\section{Convolutional DRAW}



In this section we describe the details of a single-layer version of the algorithm. Convolutional DRAW contains the following variables: input $x$, reconstruction $r$, reconstruction error $\err$, the state of the encoder recurrent net $h^e$, the state of the decoder recurrent net $h^d$ and latent variable $z$. The variables $h^e$, $h^d$ and $r$ are recurrent (passed between different time steps) and are initialized with learned biases. Then at each time step $t \in \{1, T\}$, convolutional DRAW performs the following updates:
\begin{eqnarray}
\err &=& x - r \\
h^e &=& \mbox{Rnn} (x, \err, h^e, h^d) \\
\label{eq:qz}
z &\sim& q = q(z|h^e) \\
\label{eq:pz}
p &=& p(z|h^d) \\
h^d &=& \mbox{Rnn} (z, h^d, r) \\
r &=& r + Wh^d \\
L^z_t &=& KL(q | p)
\label{eq:Lz1L}
\end{eqnarray}



where $W$ denotes a convolution and Rnn denotes a recurrent network. We use LSTM \citep{hochreiter1997long} with convolutional operators instead of the usual linear ones. The final value of $r_{final} = r_T$ contains the parameters of the input distribution. For binary images we use the Bernoulli distribution. For natural images we use the Gaussian distribution with mean and log variance given by splitting the vector $r_T$ to obtain the input cost $L^x$ and the total cost $L$:
\begin{eqnarray}
\label{eq:qx}
q^x &=& \mathcal{U}(x-s/2, x+s/2)\\
p^x &=& \mathcal{N}(r^{\mu}_T, \exp(r^{\alpha}_T)) \\
\label{eq:xKL}
L^x &=& \log(q^x/p^x)\\
L &=& \beta L^x + \sum_{t=1}^T L^z_t
\label{eq:KL1L}
\end{eqnarray} 
where the handling of real valued-ness of the inputs (\ref{eq:qx}, \ref{eq:xKL}) is explained below, and $\beta=1$ being the standard setting. The algorithm is schematically illustrated in the first layer of Figure~\ref{diagram}. The network is trained by calculating the gradient of this loss and using a stochastic gradient descent algorithm. Stochastic back-propagation through a sampling function is done as in variational auto-encoders \citep{kingma2013auto,rezende2014stochastic}. Both the approximate posterior $q$ and the prior $p$ are Gaussian, with mean and log variance being linear functions of $h^e$ or $h^d$, respectively. 

Let us discuss how we handle the input distribution for natural images. Each pixel (per color) is one of 256 values. We could use a soft-max distribution to model it. This would result in a rather large output vector at every time step and also does not take advantage of the underlying real valued-ness of intensities and therefore we opted for a Gaussian distribution. However this still needs to be converted to a discrete distribution over 256 values to calculate the negative likelihood loss $L^x = -\log p(x|z)$. Instead of this, we add uniform noise to the input with width corresponding to the spacing between discrete values and calculate $L^x = -\log p(x|z)/q^0(x)$ where $q^0(x) = \mathcal{U}(x-s/2, x+s/2)$ with $s = 1/256$ if inputs are scaled to the interval $[0, 1]$.

\subsection{Multi-layer Architectures}

Next we explain how we can stack convolutional DRAW with a two layer example. The first layer is the same as the one just introduced. The second layer has the same structure: recurrent encoder, recurrent decoder and a stochastic layer. The input to the second layer is the mean $\mu$ of the approximate posterior of the first layer. The output of the second layer biases the prior of the latent variable of the first layer and is also passed as input into the first layer decoder recurrent net. This is illustrated in Figure~\ref{diagram}. We don't use any reconstruction or error in the second layer. 

Here we describe a given computational step in detail. Indices $1$ and $2$ denote the variables of layers $1$ and $2$, respectively. In addition, let $\mu_1(q_1)$ be the mean of $q_1$. Then, the update at a given time step is given by
\begin{eqnarray}
\err &=& x - r \\
h^e_1 &=& \mbox{Rnn} (x, \err, h^e_1, h^d_1) \\
z_1 &\sim& q_1 = q_1(z_1|h^e_1) \\
h^e_2 &=& \mbox{Rnn} (\mu_1(q_1), h^e_2, h^d_2) \\
z_2 &\sim& q_2 = q_2(z_2|h^e_2) \\
p_2 &=& p(z_2|h^d_2) \\
h^d_2 &=& \mbox{Rnn} (z_2, h^d_2) \\
p_1 &=& p(z_1|h^d_1, h^d_2) \\
h^d_1 &=& \mbox{Rnn} (z_1, h^d_1, h^d_2, r) \\
r &=& r + Wh^d_1 \\
L^z_t &=& KL(q_1 | p_1)+KL(q_2|p_2)
\label{eq:Lz2L}
\end{eqnarray}
Systems with more layers can be built analogously. 

\begin{figure}[h!] 
\vspace{-.0cm}
\begin{center}
\begin{minipage}{0.5\textwidth}
\includegraphics[width=0.95\textwidth]{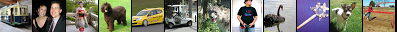}
\includegraphics[width=0.95\textwidth]{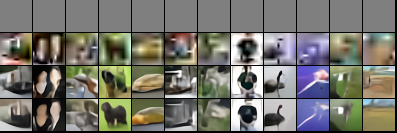}
\includegraphics[width=0.95\textwidth]{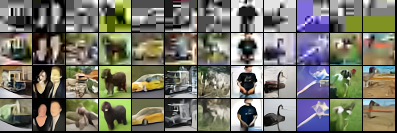}
\includegraphics[width=0.95\textwidth]{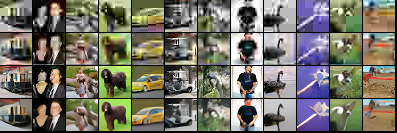}
\includegraphics[width=0.95\textwidth]{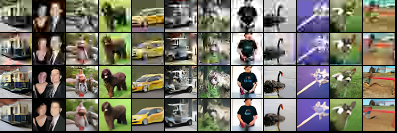}
\includegraphics[width=0.95\textwidth]{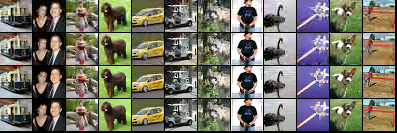}
\includegraphics[width=0.95\textwidth]{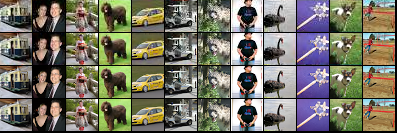}
\caption{\textbf{Lossy Compression.}
Example images for various methods and levels of compression. Top block: original images. Each subsequent block has four rows corresponding to four methods of compression: JPEG, JPEG2000, convolutional DRAW with full prior variance for generation and convolutional DRAW with zero prior variance. Each block corresponds to a different compression level; from top to bottom, average number of bits per input dimension are: 0.05, 0.1, 0.15, 0.2, 0.4, 0.8 (bits per image: 153, 307, 460, 614, 1228, 2457). In the first block, JPEG was left gray because it does not compress to this level. Images are of size $32\times32$. See appendix for $64\times64$ images. 
}
\vspace{-0.6cm}
\label{fig:compression}
\end{minipage}
\end{center}
\vspace{-.0cm}
\end{figure}

\section{Compression}

Here we show how one can turn variational auto-encoders including convolutional DRAW into compression algorithms. We have not built the actual compressor, however, as we explain, we have strong reasons to believe it would perform as well as calculated here. Two basic approaches exist. The first one is less convenient because it needs to add extra data to the bitstream when compressing an image but has essentially a guaranteed compression rate. The other one may require some experimentation but is expected to yield a similar compression rate and can be used on a given image without needing extra data.


The underlying compression mechanism for all cases is arithmetic coding \citep{witten1987arithmetic}. Arithmetic coding takes as input a sequence of discrete variables $x_1, \ldots, x_t$ and a set of probabilities $p(x_t|x_1, \ldots, x_{t-1})$ that predict the variable at time $t$ from previous variables. It then compresses this sequence to $L = -\sum_t \log_2(p(x_t|x_1, \ldots, x_{t-1}))$ bits plus a constant of order one.

Compressing inputs using variational auto-encoders proceeds as follows: discretize each latent variable in each layer using the width of $q$ (eq. \ref{eq:qz}), treat the resulting variables as a sequence with predictions $p$ (eq. \ref{eq:pz}) and compress using arithmetic coding.

For this to work as explained, several things are needed. First, the discretization should be independent of the input. This can be achieved by training the network with the variance of $q$ being a learned constant that does not depend on the input. We found that this does not have much effect on the likelihood. Second, one should evaluate the log likelihood using this discretization. One has to decide on the exact manner in which $p$ should be computed for each discrete value. Significant tuning might be required here, for the obtained likelihoods to be as good as the ones obtained with sampling. However this is likely to be fruitful since there exists a second, less convenient way to compress that is guaranteed to achieve this rate.

This second approach uses bits-back coding \citep{hinton1993keeping}. We explain only the basic idea here. We discretize the latents down to a very high level of precision and use $p$ to transmit the information. Because the discretization precision is high, the probabilities for discrete values are easily assigned. That will preserve the information but it will cost many bits, namely $-\log_2(p^d)$ where $p^d$ is a probability under that discretization. Now, instead of sampling from the approximate posterior $q$ when encoding an input, we encode $-\log_2 q(z)$ bits of other information into the choice of $z$, that we also want to transmit. When $z$ is recovered at the receiving end, both the information about the current input and the other information is recovered and thus the information needed to encode the current input is $-\log_2(p) + \log_2(q) = -\log_2(p/q)$. 
The expectation of this quantity is the KL-divergence in~(\ref{eq:Lz1L}), which therefore measures the amount of information stored in a given latent layer.
The disadvantage of this approach is that we cannot encode a given input without also having some other information we want to transmit. However, this coding scheme works even if the variance of the approximate posterior is dependent on the input.

\section{Results}

For natural images, all models except otherwise specified were single-layer, with $n_t=32$, a kernel size of $5\times5$, and stride 2 convolutions between input layers and hidden layers with $12$ latent feature maps.
We trained the models on Cifar-10 and ImageNet with $320$ and $160$ LSTM feature maps respectively. We use the version of ImageNet presented in \citep{oord2016pixel} that will soon be released as a standard dataset. We train the network with the Adam algorithm \citep{kingma2014adam} with learning rate $5 \times 10^{-4}$. Occasionally, we find that the cost suddenly increases dramatically. This is probably due to the Gaussian nature of the distribution, when a given variable is produced too far from the mean relative to sigma. We observed this happening approximately once per run. To be able to keep training we store older parameters, detect such jumps and revert to the old parameters when they occur. The network then just keeps training as if nothing had happened.

\subsection{Modeling Quality}


\begin{figure}[t]
\begin{center}
\begin{minipage}{0.5\textwidth}
\includegraphics[width=0.99\textwidth]{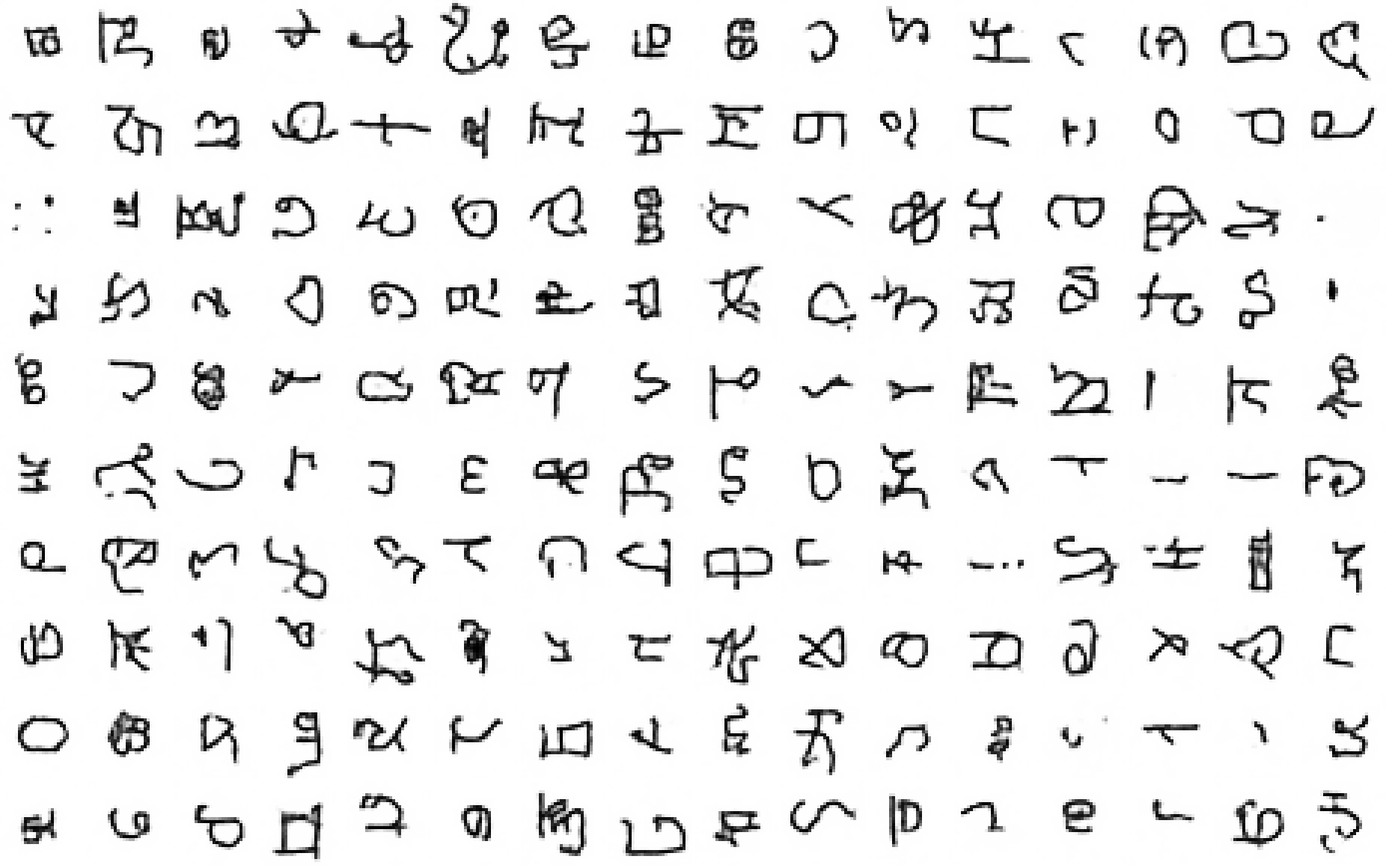}
\caption{\textbf{Generated samples for Omniglot}. 
}
\label{fig:samplesOmniglot}
\end{minipage}
\end{center}
\vspace{-0.75cm}
\end{figure}

\begin{figure}[t]
\vspace{-.0cm}
\begin{center}
\begin{minipage}{0.5\textwidth}
\includegraphics[width=0.99\textwidth]{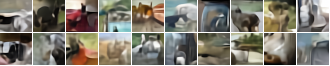}
\includegraphics[width=0.99\textwidth]{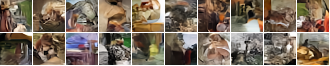}
\includegraphics[width=0.99\textwidth]{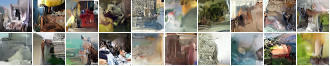}
\includegraphics[width=0.99\textwidth]{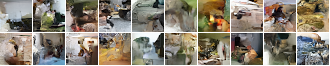}
\includegraphics[width=0.99\textwidth]{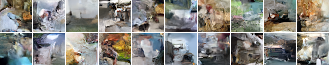}
\caption{\textbf{Generated samples for different input cost scales}. Convolutional DRAW is trained on $32 \times 32$ ImageNet. The scale of the input cost $\beta$ in (\ref{eq:KL1L}) is \{0.2, 0.4, 0.6, 0.8, 1\} for each respective block of two rows, with standard maximum likelihood being $\beta=1$. For smaller values of $\beta$ the network is less compelled to explain very fine details of images and produces cleaner larger structures. 
}
\label{fig:samples}
\end{minipage}
\end{center}
\end{figure}

\begin{figure}[h!]
\vspace{-0.3cm}
\begin{center}
\begin{minipage}{0.5\textwidth}
\includegraphics[width=0.99\textwidth]{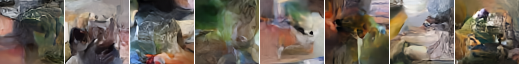}
\includegraphics[width=0.99\linewidth]{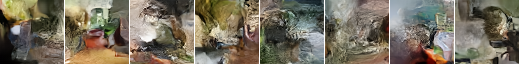}
\includegraphics[width=0.99\textwidth]{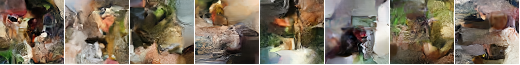}
\includegraphics[width=0.99\textwidth]{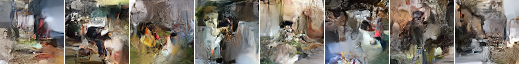}
\includegraphics[width=0.99\textwidth]{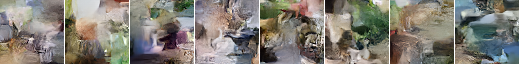}
\caption{\textbf{Generated samples for different input cost scales on $\bf{64 \times 64}$ ImageNet}. The input cost scale $\beta$ is $\{0.4, 0.5, 0.6, 0.8, 1\}$ for each row respectively. 
}
\label{fig:samples64}
\vspace{-0.65cm}
\end{minipage}
\end{center}
\end{figure}

\subsubsection{Omniglot}

The recently introduced Omniglot dataset \cite{lake2015human} is comprised of $1628$ character classes drawn from multiple alphabets with just $20$ samples per class. Referred to by many as the `inverse of MNIST', it was designed to study conceptual representations and generative models in a low-data regime. Table~\ref{table:omniglot} shows likelihoods of different models compared to ours. For our model, we only calculate the upper bound (variational bound) and therefore the true likelihood is actually better. Samples generated by the model are shown in Figure~\ref{fig:samplesOmniglot}.

\subsubsection{Cifar-10}

Table \ref{table:cifar10} shows likelihoods of different models on Cifar-10. We see that our method outperforms previous methods except for the just released Pixel RNN model of \citep{oord2016pixel}. As mentioned, the advantage of our model compared to such auto-regressive models is that it is a latent variable model that can be used for representation learning and lossy compression. At the same time, the two approaches are orthogonal and can be combined, for example by feeding the output of convolutional DRAW into the recurrent network of Pixel RNN.


We also report the likelihood for a (non-recurrent) variational auto-encoder and standard DRAW. For the variational auto-encoder we tested architectures with multiple layers, both deterministic and stochastic but with standard functional forms, and this was the best result that we obtained. Convolutional DRAW performs significantly better.


\subsubsection{ImageNet}

Additionally, we trained on the version of ImageNet prepared in \citep{oord2016pixel} which was created with the aim of making a standardized dataset to test generative models. The results are in Table \ref{table:imageNet}. Note that being a new dataset, no other methods have been reported on it.

In Figure~\ref{fig:samples} and Figure~\ref{fig:samples64} we show generations from the model. We trained networks with varying input cost scales as explained in the next section. The generations are sharp and contain many details, unlike previous versions of variational auto-encoder that tend to generate blurry images.

\begin{table}[h]
\vspace{-0.3cm}
\caption{Test set performance of different models on $28\times28$ Omniglot in \emph{nats}.}
\vspace{0.4cm}
\centering
	\begin{tabular}{lcc}
		\hline
		\textbf{Model} & \textbf{NLL Test}  \\ 
		\hline
         VAE (2 layer, 5 samples) & 106.31 \\
         IWAE (2 layer, 50 samples) & 103.38 \\
         RBM (500 hidden) & 100.46 \\
		\hline
        DRAW & $<$ 96.5\\
		\hline
        Conv DRAW & $<$ 91.0\\
	   	\hline
	\end{tabular}
\label{table:omniglot}
\end{table}

\begin{table}[h]
\vspace{-0.3cm}
\caption{Test set performance of different models on CIFAR-10 in \emph{bits/dim}. For our models we give the training performance in brackets. [1] \cite{dinh2014nice}, [2] \cite{sohl2015deep}, [3] \cite{van2014factoring}, [4] \citep{oord2016pixel}.}
\vspace{0.4cm}
\centering
	\begin{tabular}{lcc}
		\hline
		\textbf{Model} & \textbf{NLL Test (Train)}  \\ 
		\hline
		Uniform Distribution & 8.00 \\ 
		Multivariate Gaussian & 4.70 \\ 
		NICE [1] & 4.48 \\ 
		Deep Diffusion [2] & 4.20 \\ 
		Deep GMMs [3] & 4.00 \\ 
		Pixel RNN [4] & 3.00 (2.93) \\
		\hline
		Deep VAE & $<$ 4.54 \\
		DRAW  & $<$ 4.13 \\
		\hline
		Conv DRAW & $<$ 3.58 (3.57) \\
	   	\hline
	\end{tabular}
\label{table:cifar10}
\vspace{-0.3cm}
\end{table}

\begin{table}[h]
\vspace{-0.3cm}
\caption{Performance of different models on ImageNet in \emph{bits/dim}. Note that the Pixel RNN method has just been released and no other methods have been tested on this dataset.}
\vspace{0.4cm}
\centering
	\begin{tabular}{lcc}
		\hline
		\textbf{$32 \times 32$} & \textbf{Method} & \textbf{NLL Validation (Train)}  \\ 
		\hline
		& Conv DRAW & $<$ 4.40 (4.35) \\
		& Pixel RNN & 3.86 (3.83) \\ 
		\hline
		\textbf{$64 \times 64$} & \textbf{Method} & \textbf{NLL Validation (Train)}  \\ 
		\hline
		& Conv DRAW & $<$ 4.10 (4.04) \\
		& Pixel RNN & 3.63 (3.57) \\ 
	   	\hline
	\end{tabular}
\label{table:imageNet}
\end{table}

\subsection{Input Cost Scaling}
\label{sec:inputScale}

Each pixel (and color channel) of the data consists of 256 values, and as such, likelihood and lossless compression are well defined. When compressing the image there is much to be gained in capturing precise correlations between nearby pixels. There are a lot more bits in these low level details than in the  higher level structure that we are actually interested in when learning higher level representations. The network might focus on these details, ignoring higher level structure. 

One way to make it focus less on the details is to scale down the cost of the input relative to the latents, that is, setting $\beta<1$ in (\ref{eq:KL1L}). Generations for different cost scalings are shown in Figure \ref{fig:samples}, with the original objective being scale $\beta = 1$. We see that lower scales indeed have a `cleaner' high level structure. Scale 1 contains a lot of information at the precise pixel values and the network tries to capture that, while not being good enough to properly align details and produce realistic patterns. 
This might be simply a matter of scaling, making layers larger, networks deeper, using more iterations, or using better functional forms. 



\subsection{The Dependence on Computational Depth}

\begin{figure}[t]
\begin{center}
\vspace{-0.2cm}
\begin{minipage}{0.5\textwidth}
\includegraphics[width=0.49\textwidth]{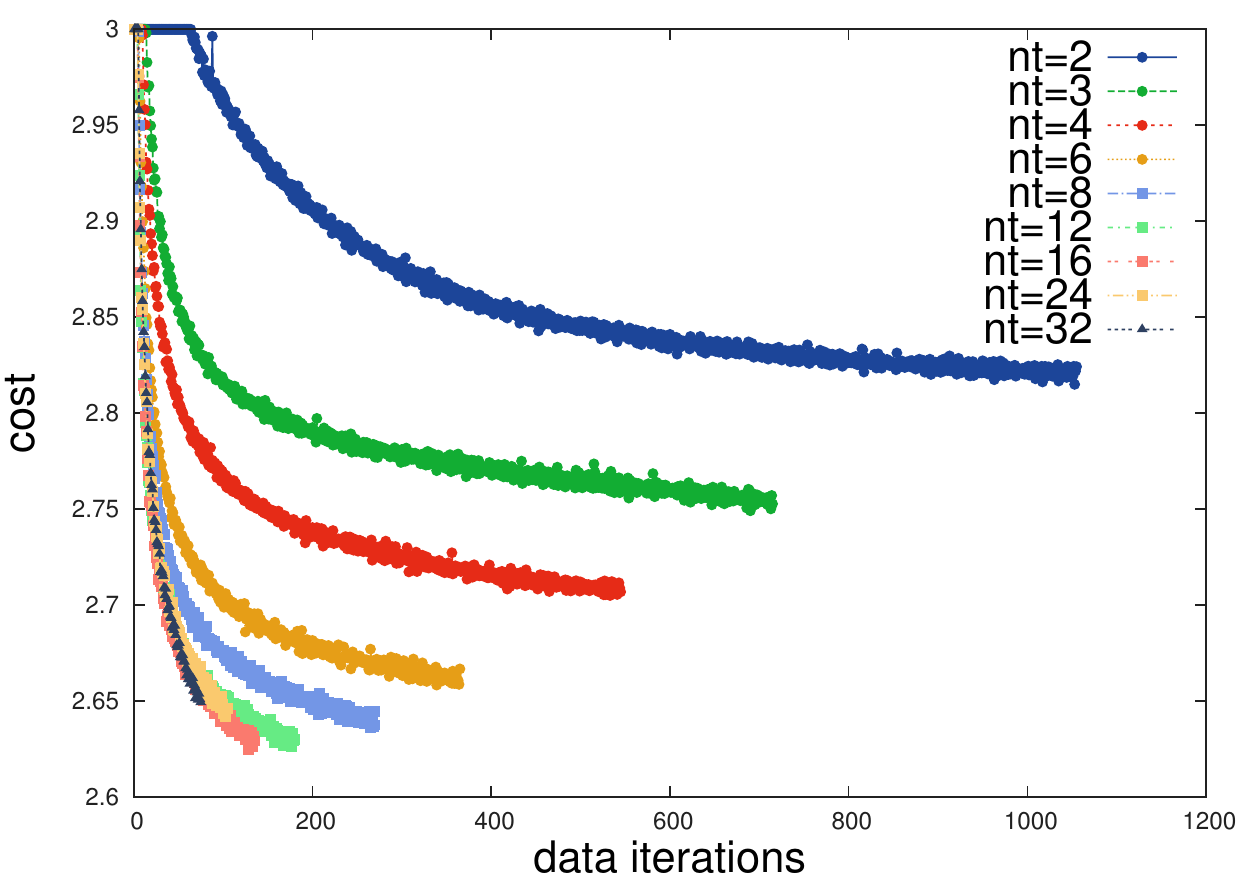}
\includegraphics[width=0.49\linewidth]{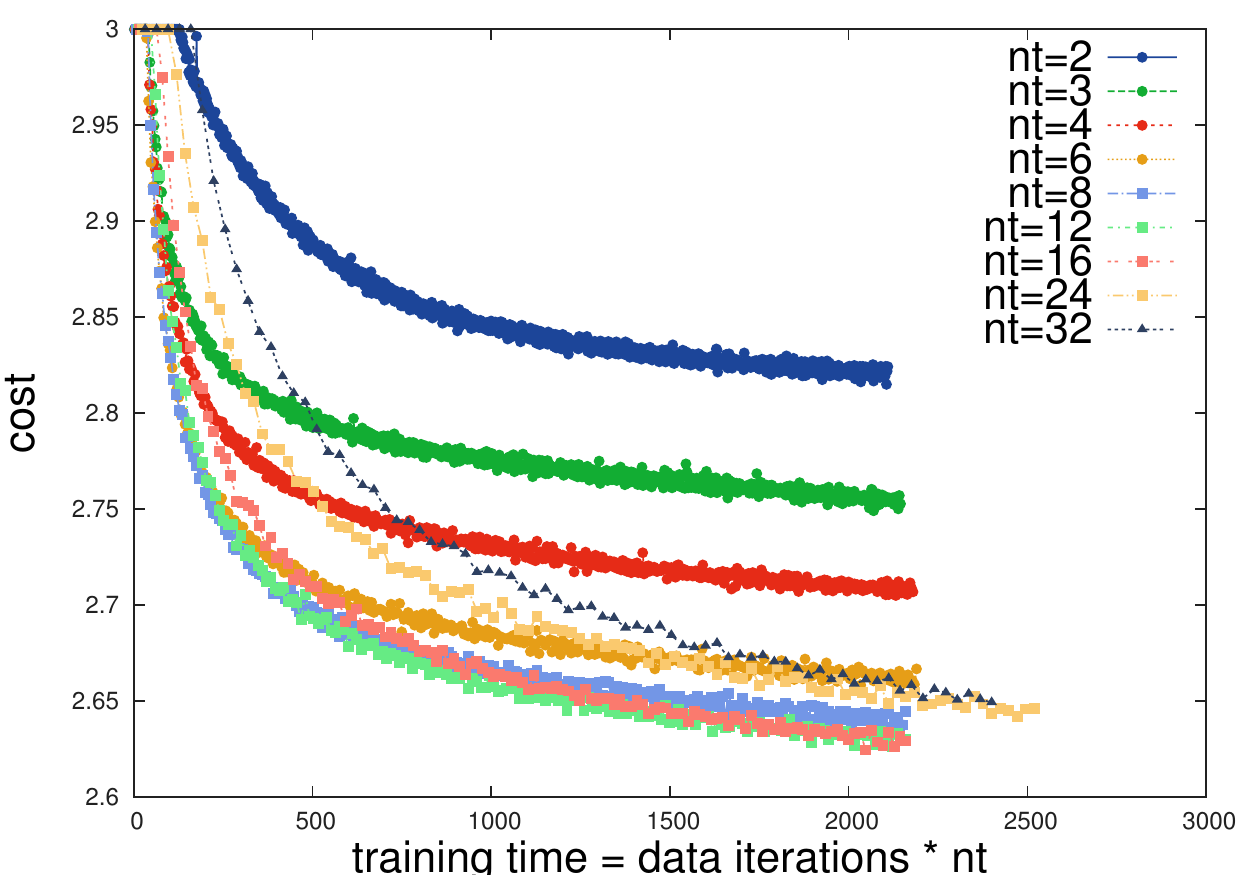}
\caption{\textbf{Dependence of training times for different number of DRAW iterations}. Convolutional DRAW performs several recurrent steps $n_t$ for a given input image. The left graph shows training curves for different $n_t$ as a function of the number of data presentations, and the right graph displays the same as a function of real training time. We see that despite having to do several iterations, up to about $n_t<16$, DRAW does not take more wall clock time to train than DRAW with smaller $n_t$. For larger $n_t$, the training slows down, however it does eventually reach better performance than lower $n_t$. }
\vspace{-0.7cm}
\label{fig:trainTime}
\end{minipage}
\end{center}
\end{figure}

Convolutional DRAW uses many iterations and might be considered expensive. However we found that networks with a larger number of time steps train faster per data example as shown in Figure \ref{fig:trainTime} (left). To study how they train with respect to real time, we multiply the time scale of each input by the number of iterations as seen in Figure \ref{fig:trainTime} (right). We see that despite having to do several iterations, up to about $n_t<16$, convolutional DRAW does not take more wall clock time to train than the same architecture with smaller $n_t$. For larger $n_t$, the training slows down, but it does eventually reach better performance than at lower $n_t$. 

\subsection{Information Distribution}

\begin{figure}[t]
\vspace{-.0cm}
\begin{center}
\begin{minipage}{0.5\textwidth}
\includegraphics[width=0.85\textwidth]{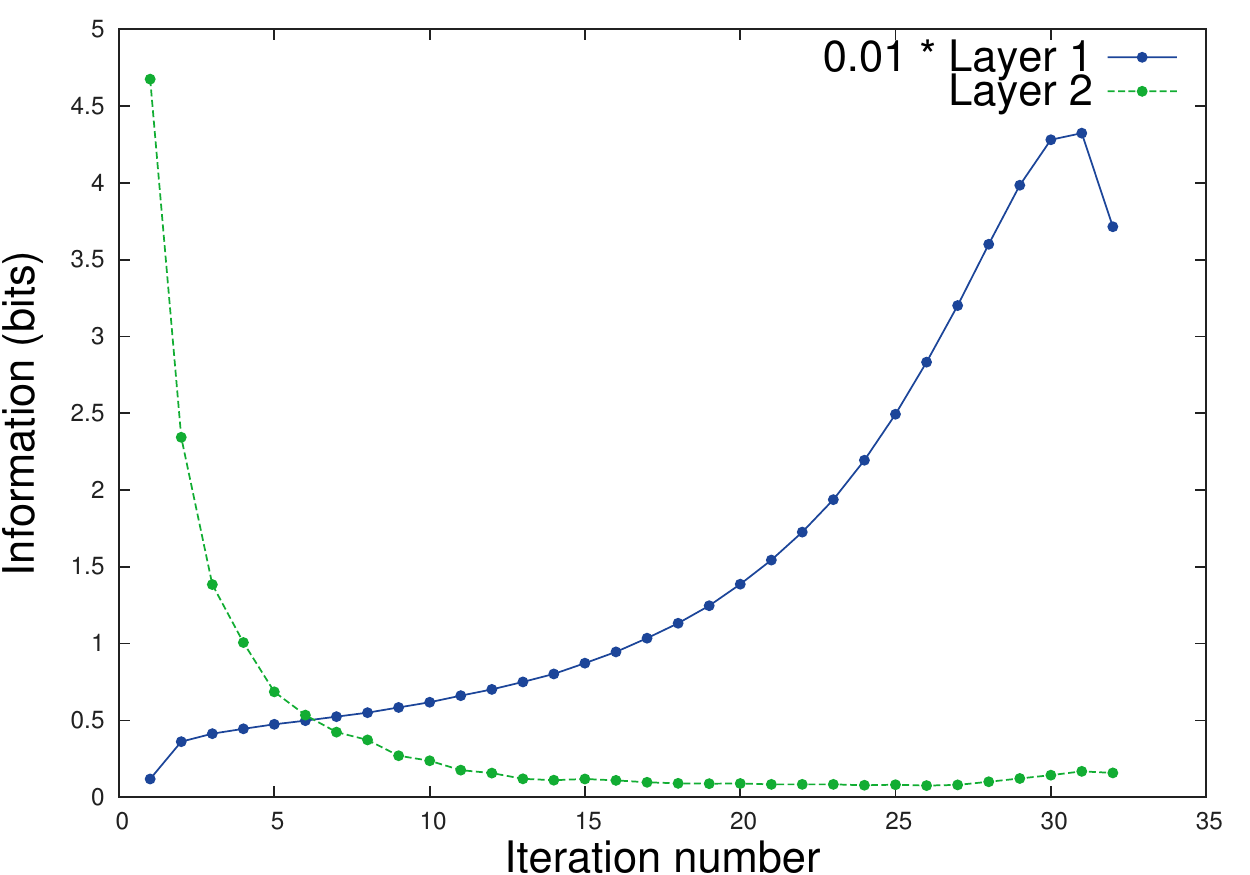}
\caption{\textbf{Amount of information at different layers and time steps.} A two-layer convolutional DRAW was trained on ImageNet, with a convolutional first layer and a fully connected second layer. The amount of information at a given layer and iteration is measured by the KL-divergence between the prior and the posterior (\ref{eq:Lz2L}). When presented with an image, first the top layer acquires information and then the second slowly increases, suggesting that the network first acquires information about `what is in the image' and subsequently encodes the details.
}
\vspace{-0.6cm}
\label{fig:LatentCost}
\end{minipage}
\end{center}
\vspace{-.0cm}
\end{figure}

We look at how much information different levels and time steps contain. This information is simply the KL divergence in (\ref{eq:Lz1L}) and (\ref{eq:Lz2L}). For a two layer system with one convolutional and one fully connected layer, this is shown in Figure \ref{fig:LatentCost}.

We see that the higher level contains information mainly at the beginning of the computation, whereas the lower layer starts with low information which then gradually increases. This is desirable from a conceptual point of view. It suggests that the network first finds out about an overall structure of the image and then explains the details contained within that structure.
Understanding the overall structure rapidly is also convenient if the algorithm needs to respond to observations in a timely manner.



\subsection{Lossy Compression}

We can compress an image with loss of information by storing only a subset of the latent variables, typically the high levels of the hierarchy. We can do this in multilayer convolutional DRAW, storing only higher levels. However we can also store only a subset of time steps, specifically a given number of time steps at the beginning, and let the network generate the rest.

The units not stored should be generated from the prior distribution (Equation~\ref{eq:pz}). However we can also generate a more likely image by lowering the variance of the prior Gaussian. We show generations with full variance in row 3 of each block of Figure \ref{fig:compression} and with variance zero in row 4 of that figure. We see that using the original variance, the network generates sharp details. Because the generative model is not perfect, the resulting images are less realistic looking as we lower the number of stored time steps. For zero variance we see that the network starts with rough details making a smooth image and then refines it with more time steps. All these generations are produced with a single-layer convolutional DRAW, and thus, despite being single-layer, it achieves some level of `conceptual compression' by first capturing the global structure of the image and then focusing on details.


There is another dimension we can vary for lossy compression -- the input scale introduced in subsection \ref{sec:inputScale}. Even if we store all the latent variables (but not the input bits), the reconstructed images will get less detailed as we scale down the input cost. 

To build a really good compressor, at each compression rate, we need to find which of the networks, input scales and number of time steps would produce visually good images. For several compression levels, we have looked at images produced by different methods and selected qualitatively which network gave the best looking images. We have not done this per image, just per compression level. We then display compressed images that we have not seen with this selection.

We compare our results to JPEG and JPEG2000 compression which we obtained using ImageMagick. We found however that these compressors are unable to produce reasonable results for small images ($3 \times 32 \times 32$) at high compression rates. Instead, we concatenated 100 images into one $3 \times 320 \times 320$ image, compressed that and extracted back the compressed small images. The number of bits per image reported is then the number of bits of this image divided by 100. This is actually unfair to our algorithm since any correlations between nearby images can be exploited. Nevertheless we show the comparison in Figure \ref{fig:compression}. Our algorithm shows better quality than JPEG and JPEG 2000 at all levels where a corruption is easily detectable. Note that even when our algorithm is trained on one specific image size, it can be used on arbitrarily sized images for those networks that contain only convolutional operators. 


\section{Conclusion}

In this paper, we introduced convolutional DRAW, a state-of-the-art generative model which demonstrates the potential of sequential computation and recurrent neural networks in scaling up latent variable models. During inference, the algorithm sequentially arrives at a natural stratification of information, ranging from global aspects to low-level details. An interesting feature of the method is that, when we restrict ourselves to storing just the high level latent variables, we arrive at a `conceptual compression' algorithm that rivals the quality of JPEG2000. As a generative model, it outperforms earlier latent variable models on both the Omniglot and ImageNet datasets.



\vspace{-0.2cm}
\section*{Acknowledgements} 
We thank Aaron van den Oord, Diederik Kingma and Koray Kavukcuoglu for fruitful discussions. 
 \vspace{-0.2cm}

\nocite{langley00}

\bibliography{convDrawNew}
\bibliographystyle{icml2016}
\vspace{-0.2cm}
\section*{Appendix}

\begin{figure*} 
\centering
\includegraphics[width=0.98\textwidth]{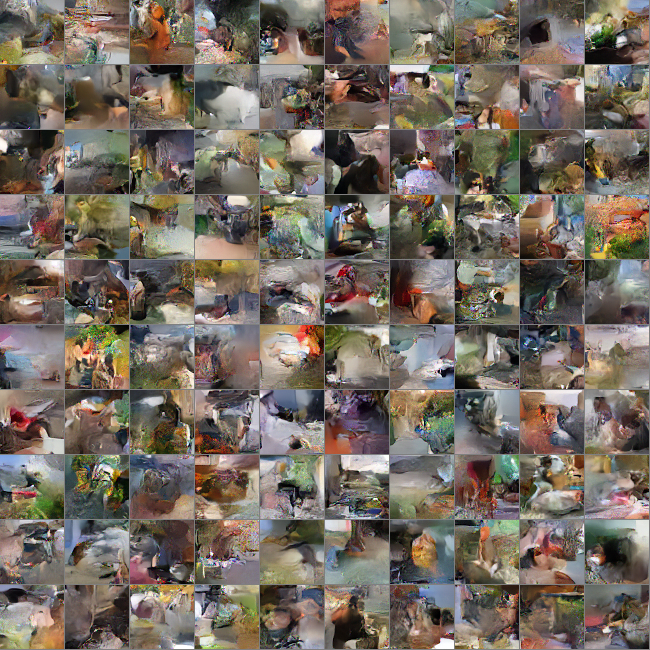}
\caption{
\textbf{Generated samples from a network trained on $64 \times 64$ ImageNet with input scaling $\beta = 0.4$}. Qualitatively asking the model to be less precise seems to lead to visually more appealing samples.}
\label{fig:gen64a}
\end{figure*}

\begin{figure*} 
\centering
\includegraphics[width=0.98\textwidth]{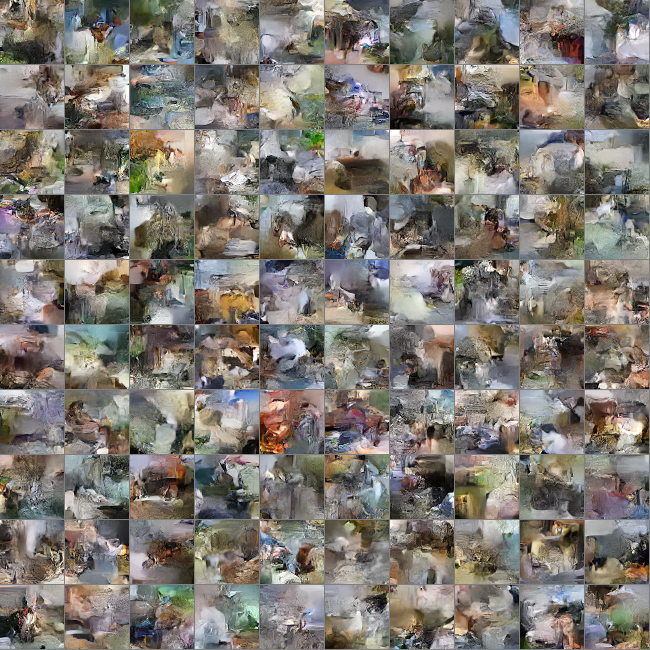}
\caption{
\textbf{Generated samples from a network trained on $64 \times 64$ ImageNet with input scaling $\beta = 1$}. For this value of $\beta$, the system dedicates a lot of resources to explain details, losing higher level coherence. As models get better, this problem might disappear. }
\label{fig:gen64b}
\end{figure*}

\begin{figure*} 
\centering
\includegraphics[width=0.95\textwidth]{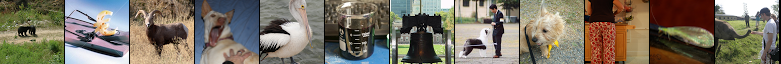}
\includegraphics[width=0.95\textwidth]{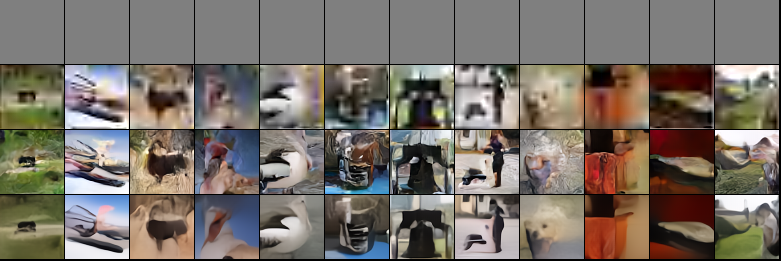}
\includegraphics[width=0.95\textwidth]{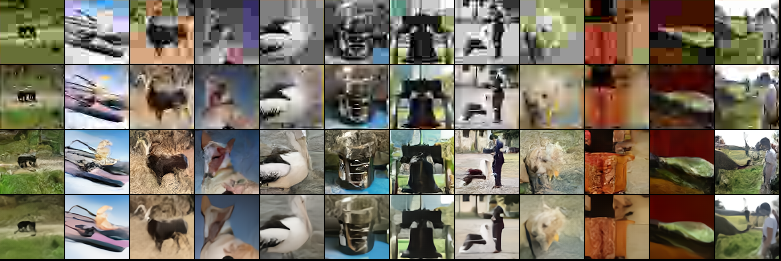}
\includegraphics[width=0.95\textwidth]{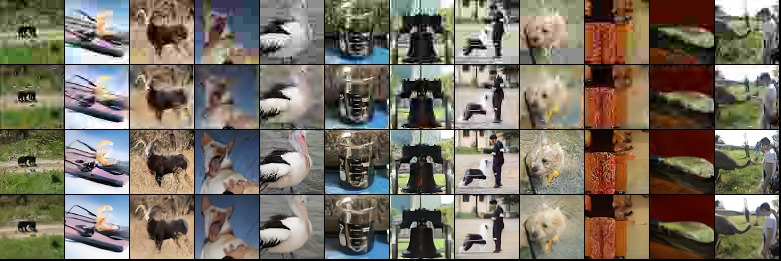}
\caption{\textbf{Lossy Compression, Part 1}
Analogous to Figure~\ref{fig:compression} of the main paper but for $64 \times 64$ inputs. Example images for various methods and amounts of compression. Top block: original images. Each subsequent block has four methods of compression: JPEG, JPEG2000, convolutional DRAW with full prior variance for generation and convolutional DRAW with zero prior variance. Different blocks correspond to different compression levels, from top to bottom with bits per input dimension: 0.05, 0.1, 0.15, 0.2, 0.4, 0.8. In the first block, JPEG was left gray because it does not compress to this level. }
\label{fig:compression64}
\end{figure*}

\begin{figure*} 
\centering
\includegraphics[width=0.95\textwidth]{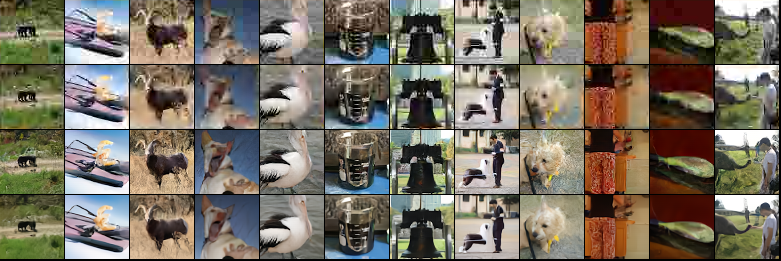}
\includegraphics[width=0.95\textwidth]{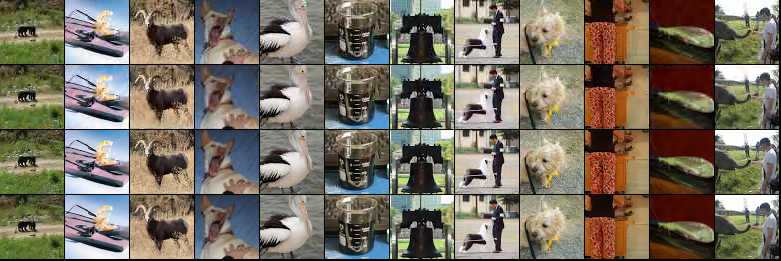}
\includegraphics[width=0.95\textwidth]{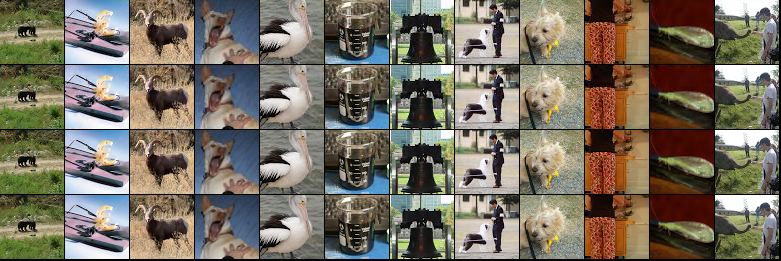}
\caption{\textbf{Lossy Compression, Part 2}}
\label{fig:compression64b}
\end{figure*}

Figures \ref{fig:gen64a}, \ref{fig:gen64b} show $32 \times 32$ image generations for input scaling $\beta = 0.4$ and $\beta = 1$ of (\ref{eq:KL1L}), while Figures \ref{fig:compression64}, \ref{fig:compression64b} show $64 \times 64$ generations, also for  $\beta = 0.4$ and $\beta = 1$.

\end{document}